\documentclass[conference]{IEEEtran}
\IEEEoverridecommandlockouts
\usepackage{cite}
\usepackage{amsmath,amssymb}
\usepackage{graphicx}
\usepackage{booktabs}
\usepackage{multirow}
\usepackage{textcomp}
\usepackage{xcolor}
\usepackage{tikz}
\usetikzlibrary{positioning,arrows.meta,decorations.pathreplacing}
\usepackage[hidelinks]{hyperref}
\usepackage{url}

\begin{document}

\title{Bengali Document Layout Analysis with Detectron2}

\author{
\IEEEauthorblockN{Md Ataullha, Mahedi Hassan Rabby, Mushfiqur Rahman, Tahsina Bintay Azam}
\IEEEauthorblockA{Department of Computer Science and Engineering\\
Shahjalal University of Science and Technology, Sylhet, Bangladesh\\
\{ataullha81, mahedi33, mushfiqur83, tahsina72\}@student.sust.edu}
}

\maketitle

\begin{abstract}
Document layout analysis (DLA) splits a page into regions such as paragraphs, text boxes, images and tables before optical character recognition is applied. For Bengali, the BaDLAD dataset now makes it possible to train and compare DLA models at scale. In this work we train Mask R-CNN models in Detectron2 on BaDLAD using three backbones (ResNet-50, ResNet-101 and ResNeXt-101 32$\times$8d, each with a feature pyramid network) and two initializations: an ImageNet-pretrained backbone, and a full model pretrained on the English PubLayNet dataset. All six models were scored with the Dice coefficient on the DL Sprint 2.0 test set (about 3{,}900 images) and hidden test set (about 9{,}100 images). ResNet-101 with ImageNet initialization obtained the highest Dice on both sets, 0.89191 and 0.89283. The five models with both scores were ranked in the same order on the two sets. PubLayNet initialization gave higher Dice for ResNet-50 and ResNeXt-101, where both initializations were trained for a similar number of iterations. For ResNet-101 the two runs had different training lengths, so the effect of initialization could not be separated. On one NVIDIA P100 GPU, the ResNet-101 pipeline processed 4.67 images per second including post-processing.
\end{abstract}


\section{Introduction}
Digitizing printed Bengali material, such as newspapers, books, government records and historical archives, requires the text and non-text regions of each page to be located before character recognition. This step is called document layout analysis (DLA). If a text box is missed at this stage it is not transcribed, and if two paragraphs are merged their reading order may be wrong.

Most deep learning work on DLA has used English datasets such as PubLayNet~\cite{zhong2019publaynet}, DocBank~\cite{li2020docbank} and DocLayNet~\cite{pfitzmann2022doclaynet}. BaDLAD~\cite{shihab2023badlad} provides a comparable resource for Bengali: 33{,}695 human-annotated pages from six domains with about 710K polygon annotations in four classes. Unlike the scientific articles in PubLayNet and DocBank, BaDLAD contains magazines, novels, notices, property deeds and historical documents, many with irregular layouts, degraded scans and many small text boxes.

BaDLAD was used in the DL Sprint 2.0 competition~\cite{dlsprint2}, and several participating teams have described their systems~\cite{datta2023mrcnn,ahmed2023yolov8,chak2023anchorfree,rahman2023mvitv2,hasan2023framework}. These reports mostly present one final configuration. In this paper we focus on Mask R-CNN~\cite{he2017maskrcnn}, a common baseline for DLA, and examine two choices: the backbone network, and whether starting from a model trained on English layouts (PubLayNet) helps on Bengali pages.

The main points of this paper are as follows.
\begin{itemize}
\item We compare six Mask R-CNN configurations (three backbones and two initializations) on BaDLAD and report the Dice of each on the test and hidden test sets of DL Sprint 2.0.
\item We analyze how training length, initialization and the choice of evaluation set relate to the observed differences, and point out which comparisons are affected by differences in training settings.
\item We describe the inference and post-processing steps used to produce competition-format masks and report the measured inference speed.
\end{itemize}
Compared with the first version of this preprint, this version adds hidden-test results, related work and the analysis in Sections~\ref{sec:results} and~\ref{sec:discussion}, and corrects the test-set size and inference speed reported there.

\section{Related Work}
\textbf{Datasets.} PubLayNet~\cite{zhong2019publaynet} contains more than 360,000 page images from biomedical articles with automatically generated annotations, and DocBank~\cite{li2020docbank} provides token-level labels for arXiv papers. Both come from scientific publishing. DocLayNet~\cite{pfitzmann2022doclaynet} contains 80{,}863 manually annotated pages from a wider range of document types. BaDLAD~\cite{shihab2023badlad} is described by its authors as the first multi-domain Bengali DLA dataset. They trained Faster R-CNN and Mask R-CNN with ResNet-50 backbones for 10{,}000 iterations and reported an mAP of 0.343 for Mask R-CNN, with 7.34 AP for the Table class. They also reported that a Faster R-CNN pretrained on PubLayNet improved table detection.

\textbf{Methods.} Mask R-CNN~\cite{he2017maskrcnn} with a feature pyramid network (FPN)~\cite{lin2017fpn} is widely used for DLA and is available in Detectron2~\cite{wu2019detectron2} and LayoutParser~\cite{shen2021layoutparser}. Transformer-based models pretrained on document images, such as DiT~\cite{li2022dit} and LayoutLMv3~\cite{huang2022layoutlmv3}, have reported strong results on PubLayNet. Single-stage detectors of the YOLO family~\cite{jocher2023yolov8} are also used, mainly because of their speed.

\textbf{DL Sprint 2.0 systems.} Datta et~al.~\cite{datta2023mrcnn} trained Mask R-CNN R-50 over several consecutive sessions and found that PubLayNet weights helped in the first 10{,}000 iterations but not after 22{,}000 iterations. Ahmed et~al.~\cite{ahmed2023yolov8} and Chak and Rahman~\cite{chak2023anchorfree} used ensembles of YOLOv8 models with additional post-processing. Rahman and Azad~\cite{rahman2023mvitv2} used an MViTv2-B backbone~\cite{li2022mvitv2} with Cascade Mask R-CNN~\cite{cai2018cascade} and obtained the highest Dice in the competition. Hasan et~al.~\cite{hasan2023framework} compared Detectron2, YOLOv8 and SAM on BaDLAD. Our work differs from these in that it varies the backbone and the initialization within one architecture and reports results on both evaluation sets.

\begin{figure*}[t]
\centering
\begin{tikzpicture}[node distance=4.5mm, every node/.style={font=\footnotesize, align=center},
  box/.style={draw, rounded corners=2pt, minimum height=10mm, text width=16.5mm, inner sep=2pt},
  arr/.style={-{Stealth[length=1.8mm]}, semithick}]
\node[box] (img) {Page image};
\node[box, right=of img] (bb) {Backbone\\+ FPN};
\node[box, right=of bb] (rpn) {RPN +\\ROI heads};
\node[box, right=of rpn] (inst) {Instances\\(score $\geq 0.5$)};
\node[box, right=of inst] (acc) {Keep score\\$\geq \tau$};
\node[box, right=of acc] (uni) {Union per\\class};
\node[box, right=of uni] (rle) {RLE per\\class};
\node[box, right=of rle] (csv) {Submission\\rows};
\foreach \a/\b in {img/bb, bb/rpn, rpn/inst, inst/acc, acc/uni, uni/rle, rle/csv} \draw[arr] (\a) -- (\b);
\draw[decorate, decoration={brace, amplitude=4pt, mirror}] ([yshift=-1.5mm]bb.south west) -- ([yshift=-1.5mm]inst.south east)
  node[midway, below=2.5mm, font=\scriptsize] {Mask R-CNN (Detectron2)};
\draw[decorate, decoration={brace, amplitude=4pt, mirror}] ([yshift=-1.5mm]acc.south west) -- ([yshift=-1.5mm]csv.south east)
  node[midway, below=2.5mm, font=\scriptsize] {Post-processing};
\end{tikzpicture}
\caption{Inference pipeline. Mask R-CNN predicts instance masks. Instances with confidence of at least $\tau$ are merged into one binary mask per class, and each class mask is run-length encoded in the competition format.}
\label{fig:pipeline}
\end{figure*}

\section{Dataset and Evaluation}
\label{sec:data}
\textbf{BaDLAD.} The dataset~\cite{shihab2023badlad} contains books and magazines, government documents, liberation war documents, old and new newspapers, and property deeds. Each page is annotated with polygons for four classes: paragraph, text box, image and table. The classes are very unbalanced. In the BaDLAD training split, paragraphs make up 49.3\% and text boxes 48.0\% of the 425{,}115 annotations, while images make up 2.4\% and tables only 0.32\% (1{,}353 instances)~\cite{shihab2023badlad}.

\textbf{Data splits.} DL Sprint 2.0 provided 20{,}365 labeled training images in COCO format~\cite{lin2014coco} and 13{,}000 evaluation images without labels. Table~\ref{tab:splits} lists the splits used in this paper. For the ResNet-50 and ResNeXt-101 runs we held out 5\% and 1\% of the training images, respectively, as a validation set for monitoring training; the ResNet-101 runs used all training images. We do not report validation scores. Kaggle divides the 13{,}000 evaluation images into a test set of about 30\% (about 3{,}900 images), which is scored during the competition (public leaderboard), and a hidden test set of the remaining 70\% (about 9{,}100 images), which is scored after the competition ends (private leaderboard). Kaggle gives these proportions only approximately, so the image counts are approximate. Since models are chosen during the competition by their test-set score, the hidden test set gives a less biased estimate, and we report both.

\begin{table}[t]
\centering
\caption{Data splits. Test and hidden test counts follow Kaggle's approximate 30/70 division of the 13{,}000 evaluation images.}
\label{tab:splits}
\setlength{\tabcolsep}{3pt}
\begin{tabular}{lrp{0.47\columnwidth}}
\toprule
Split & Images & Use\\
\midrule
Train & 20{,}365 & Training (minus the validation subset)\\
Validation & $\approx$1{,}018 / 204 / 0 & Held out from training data for monitoring (R-50 / X-101 / R-101)\\
Test & $\approx$3{,}900 & Scored during the competition; used for model selection\\
Hidden test & $\approx$9{,}100 & Scored after the competition\\
\bottomrule
\end{tabular}
\end{table}

\textbf{Metric.} For each evaluation image and each class, a submission gives one binary mask, encoded with the organizers' run-length encoding (RLE), which is column-major and one-indexed. Bounding boxes are not evaluated. Predicted masks are compared with the ground truth using the Dice coefficient~\cite{dice1945},
\begin{equation}
\mathrm{Dice}(P,G)=\frac{2\,|P\cap G|}{|P|+|G|},
\end{equation}
where $P$ and $G$ are the predicted and ground-truth pixel sets. The competition page does not state how the scores are averaged over images and classes, or how an image with an empty prediction and an empty ground truth is scored. We therefore report the scores computed by Kaggle as they are.

\section{Method}
\label{sec:method}

\subsection{Architecture}
All models are Mask R-CNN~\cite{he2017maskrcnn} with an FPN~\cite{lin2017fpn}, built from the Detectron2~\cite{wu2019detectron2} model-zoo configurations \texttt{mask\_rcnn\_\{R\_50, R\_101, X\_101\_32x8d\}\_FPN\_3x}. The backbones are ResNet-50, ResNet-101~\cite{he2016resnet} and ResNeXt-101 32$\times$8d~\cite{xie2017resnext}. The box and mask heads predict four classes.

\subsection{Initialization}
We compare two initializations. With \emph{ImageNet} initialization, the backbone starts from the ImageNet classification weights given in the model-zoo configuration, and the detection heads start from random weights; no COCO detector weights are used. With \emph{PubLayNet} initialization, the whole network starts from a Mask R-CNN with the same backbone trained on PubLayNet~\cite{zhong2019publaynet}, an English layout dataset with five classes, and the class-specific output layers are re-initialized for the four BaDLAD classes.

\subsection{Training}
We used the default Detectron2 trainer with automatic mixed precision. The ResNet-101 runs used 8 images per batch, 128 sampled regions per image in the ROI heads, and the default Detectron2 data augmentation for these configurations (multi-scale resizing of the shorter side and random horizontal flipping). Table~\ref{tab:config} lists the learning rate, warm-up length and number of iterations of each run; these settings were not the same for all runs. The learning rate stays constant after the linear warm-up in every run, because the decay steps of the 3$\times$ schedule (210K and 250K iterations) are never reached. The ResNet-101 run with ImageNet initialization was trained over several Kaggle sessions, each starting from the weights saved at the end of the previous one, for 64{,}000 iterations in total.

\subsection{Inference and Post-processing}
Fig.~\ref{fig:pipeline} shows the inference steps. Detectron2 removes detections with confidence below 0.5 and pastes each remaining instance mask, thresholded at 0.5, onto the full-size image. We then keep only instances whose confidence is at least $\tau$, take the union of the kept masks of each class, and encode each of the four class masks with the competition RLE. We used $\tau=0.6$ for all runs except ResNet-50 with PubLayNet initialization, for which $\tau=0.7$. We did not tune $\tau$ on held-out data.

\begin{table}[t]
\centering
\caption{Training settings. Epochs are computed as iterations $\times$ 8 / 20{,}365, assuming a batch size of 8 for all runs.}
\label{tab:config}
\setlength{\tabcolsep}{3.2pt}
\begin{tabular}{llccrrc}
\toprule
Backbone & Init. & LR & Warm-up & Iters & Epochs & Train split\\
\midrule
\multirow{2}{*}{R-50}  & ImageNet  & 0.0015 & 500  & 17{,}000 & 6.7  & 0.95\\
                       & PubLayNet & 0.001  & 1800 & 18{,}000 & 7.1  & 0.95\\ \addlinespace[2pt]
\multirow{2}{*}{R-101} & ImageNet  & 0.001  & 5    & 64{,}000 & 25.1 & 1.00\\
                       & PubLayNet & 0.001  & 5    & 25{,}000 & 9.8  & 1.00\\ \addlinespace[2pt]
\multirow{2}{*}{X-101} & ImageNet  & 0.001  & 5    & 25{,}000 & 9.8  & 0.99\\
                       & PubLayNet & 0.001  & 1000 & 28{,}255 & 11.1 & 0.99\\
\bottomrule
\end{tabular}
\end{table}

\section{Results}
\label{sec:results}

\begin{table}[t]
\centering
\caption{Dice on the test ($\approx$3{,}900 images) and hidden test ($\approx$9{,}100 images) sets, computed by Kaggle. The best value in each column is in bold.}
\label{tab:results}
\setlength{\tabcolsep}{5pt}
\begin{tabular}{llccc}
\toprule
Backbone & Init. & Test & Hidden test & Hidden $-$ Test\\
\midrule
\multirow{2}{*}{R-50}  & ImageNet  & 0.87769 & 0.87851 & $+$0.00082\\
                       & PubLayNet & 0.88099 & 0.88001 & $-$0.00098\\ \addlinespace[2pt]
\multirow{2}{*}{R-101} & ImageNet  & \textbf{0.89191} & \textbf{0.89283} & $+$0.00092\\
                       & PubLayNet & 0.89082 & 0.89216$^{\dagger}$ & $+$0.00134\\ \addlinespace[2pt]
\multirow{2}{*}{X-101} & ImageNet  & 0.87803 & n/a$^{\ddagger}$ & --\\
                       & PubLayNet & 0.88711 & 0.88878 & $+$0.00167\\
\bottomrule
\multicolumn{5}{p{0.93\columnwidth}}{\footnotesize $^{\dagger}$From the private leaderboard entry of the team account that made this submission; it may correspond to a different submission from the same account. $^{\ddagger}$Not available.}
\end{tabular}
\end{table}

\subsection{Backbone and Initialization}
Table~\ref{tab:results} gives the Dice of the six models. ResNet-101 with ImageNet initialization has the highest Dice on both sets: 0.89191 on the test set and 0.89283 on the hidden test set. ResNet-101 with PubLayNet initialization is second, with 0.89082 and 0.89216. On the test set, both ResNet-101 models are at least 0.0037 higher than every ResNet-50 and ResNeXt-101 model.

ResNeXt-101 did not perform better than ResNet-101. With PubLayNet initialization it reached 0.88711 on the test set, which is 0.0037 lower than ResNet-101 with the same initialization and a similar number of iterations (28{,}255 and 25{,}000). Since each configuration was trained only once, we cannot tell whether this difference comes from the backbone or from run-to-run variation.

Compared with ImageNet initialization, PubLayNet initialization changed Dice by $+0.00330$ (test) and $+0.00150$ (hidden test) for ResNet-50, by $+0.00908$ (test) for ResNeXt-101, and by $-0.00109$ (test) and $-0.00067$ (hidden test) for ResNet-101. For ResNet-50 and ResNeXt-101, the two runs had similar numbers of iterations, and PubLayNet initialization gave the higher score. For ResNet-101, the ImageNet run was trained 2.56$\times$ longer (64{,}000 against 25{,}000 iterations), so its slightly higher score cannot be attributed to the initialization.

\begin{figure}[t]
\centering
\includegraphics[width=\columnwidth]{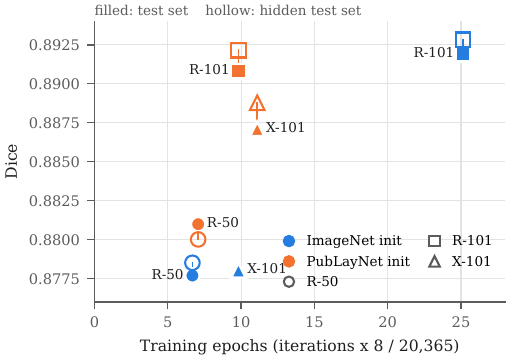}
\caption{Dice against training length. Color shows the initialization and marker shape the backbone. Filled markers are test-set scores and hollow markers hidden-test scores.}
\label{fig:epochs}
\end{figure}

\subsection{Training Length}
Fig.~\ref{fig:epochs} plots Dice against training length. Over the six runs, the Spearman rank correlation between the number of iterations and the test-set Dice is 0.81. The runs also differ in other settings, so this value only describes the six runs; it does suggest that training length may account for part of the differences we attribute to backbone or initialization.

\subsection{Test and Hidden Test Scores}
\begin{figure}[t]
\centering
\includegraphics[width=0.92\columnwidth]{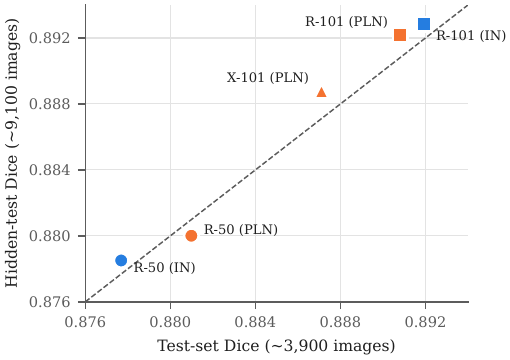}
\caption{Test versus hidden-test Dice for the five models with both scores (IN: ImageNet, PLN: PubLayNet). The dashed line is $y=x$.}
\label{fig:split}
\end{figure}

Fig.~\ref{fig:split} compares the test and hidden-test scores of the five models that have both. The models are ranked in the same order on the two sets (Spearman $\rho=1.0$). For the same model, the difference between the two scores ranged from 0.00082 to 0.00167 (mean 0.00115). If Kaggle's Dice is an equally weighted mean over images and classes, weighting the two sets by their approximate sizes gives about 0.8926 for ResNet-101/ImageNet on all 13{,}000 evaluation images.

\subsection{Other DL Sprint 2.0 Systems}
Table~\ref{tab:compare} lists the test-set Dice reported by other published DL Sprint 2.0 systems together with our best model. These systems differ in architecture, ensembling, post-processing and training length, so the table is not a controlled comparison. Our single Mask R-CNN model is 0.0035 below the two-model YOLOv8 system of Ahmed et~al.~\cite{ahmed2023yolov8} and 0.009 below the MViTv2-based system~\cite{rahman2023mvitv2}. The highest hidden-test score in the competition was 0.90396.

\begin{table}[t]
\centering
\caption{Test-set (public leaderboard) Dice reported by published DL Sprint 2.0 systems.}
\label{tab:compare}
\setlength{\tabcolsep}{3pt}
\begin{tabular}{p{0.60\columnwidth}c}
\toprule
System & Test Dice\\
\midrule
MViTv2-B + Cascade Mask R-CNN~\cite{rahman2023mvitv2} & 0.90095\\
YOLOv8m-seg, 2 models + post-proc.~\cite{ahmed2023yolov8} & 0.89537\\
YOLOv8, 5-model ensemble~\cite{chak2023anchorfree} & 0.89277\\
\textbf{Mask R-CNN R-101, single model (ours)} & \textbf{0.89191}\\
Mask R-CNN R-50, 115K iters.~\cite{datta2023mrcnn} & 0.88900$^{*}$\\
\bottomrule
\multicolumn{2}{p{0.93\columnwidth}}{\footnotesize $^{*}$Reported as a test score; the paper does not say which evaluation set.}
\end{tabular}
\end{table}

\subsection{Inference Speed}
With a batch size of 8 on one NVIDIA Tesla P100 GPU, the ResNet-101 model processed the 13{,}000 evaluation images in 46\,min 24\,s, that is, 0.214\,s or 4.67 images per second. This time includes data loading, the forward pass, pasting masks at full resolution, the per-class union and RLE encoding. The complete Kaggle notebook run, including environment setup, took 2{,}904\,s. These times cannot be compared directly with the Detectron2 model-zoo timings, which cover only the forward pass on different hardware.

\subsection{Qualitative Results}
\begin{figure*}[t]
\centering
\includegraphics[width=0.78\textwidth]{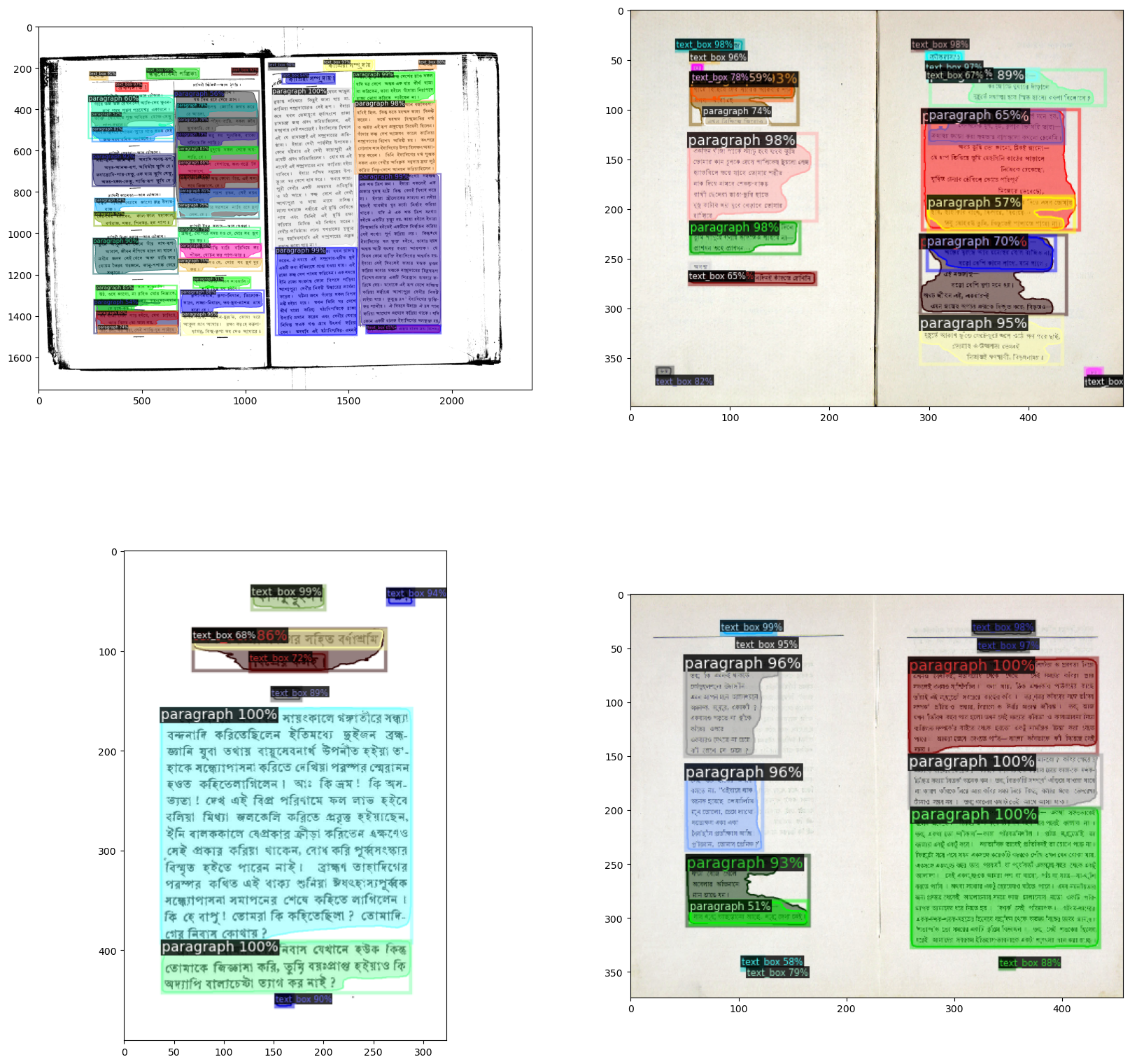}
\caption{Predictions of Mask R-CNN R-101 (ImageNet initialization) on four evaluation pages, with class labels and confidence scores.}
\label{fig:qual}
\end{figure*}

Fig.~\ref{fig:qual} shows ResNet-101 predictions on four evaluation pages. Large single-column paragraphs are segmented well. We observed three kinds of errors: several overlapping detections of the same class on one region, such as a heading detected as more than one text box; masks extending past the paragraph boundary into blank space on low-contrast scans; and broken, low-confidence detections of small text boxes on two-page spreads. Since all instances of a class are merged into one mask before scoring, overlapping detections do not lower the Dice, but broken detections and masks that extend too far do.

\section{Discussion}
\label{sec:discussion}
Among the configurations we trained, ResNet-101 gave the highest Dice on both the test and hidden test sets. Because the ranking was the same on both sets, choosing models by their test-set score did not lead to a different choice than the hidden test set would have. Where training lengths were similar (ResNet-50 and ResNeXt-101), PubLayNet initialization gave higher Dice, which agrees with the improvement in table detection reported by the BaDLAD authors~\cite{shihab2023badlad}.

The two ResNet-101 models cannot be ranked from these results. The ImageNet run was trained 2.56$\times$ longer than the PubLayNet run, and the difference between them (0.00067 to 0.00109) is smaller than the largest difference we observed between the test and hidden-test scores of a single model (0.00167). Comparing them properly would require equal training lengths, several random seeds and a paired statistical test.

For a Bengali OCR pipeline running on a single GPU, a ResNet-101 Mask R-CNN is a simple single-model option. It ran at about 4.7 images per second on a P100 including post-processing, and its Dice was 0.009 lower on the test set and 0.011 lower on the hidden test set than the best competition system.

\section{Limitations}
The six runs differ in learning rate, warm-up, number of iterations and the fraction of training data used, as well as in backbone and initialization. Each configuration was trained once with a time-based random seed, so we cannot report variance across runs. The ResNet-101 models used all training images, so they were selected using the test set; the hidden test set, however, ranks them the same way. We report only the competition's Dice score and not COCO mask AP or per-class scores, which would show performance on the rare Table class. The threshold $\tau$ was fixed rather than tuned on held-out data.

\section{Conclusion}
We trained Mask R-CNN models with three backbones and two initializations on BaDLAD and evaluated them on the test and hidden test sets of DL Sprint 2.0. A single ResNet-101 model reached a Dice of 0.89283 on the hidden test set. The models were ranked in the same order on both sets, and PubLayNet initialization gave higher Dice where training lengths were similar. Part of the remaining differences may be explained by training length. A comparison with equal training schedules, several seeds, a fixed validation set and per-class mask AP is needed to determine when pretraining on English layouts helps for Bengali documents.

\section*{Acknowledgment}
We thank the organizers of DL Sprint 2.0 and Bengali.AI for providing the BaDLAD dataset and the competition platform. An AI writing assistant (Claude, Anthropic) was used to help revise the text of this manuscript; the authors checked all content and are responsible for it.

\bibliographystyle{IEEEtran}
\bibliography{references}

\end{document}